\documentclass[10pt]{article}

\usepackage{graphicx} 
\usepackage{amsmath}
\usepackage{lipsum}
\usepackage{pdfcomment}

\usepackage{hyperref} 
\usepackage[english]{babel}
\usepackage[utf8]{inputenc}
\usepackage{amsfonts}
\usepackage{longtable}
\usepackage{booktabs}
\usepackage{float} 
\usepackage{geometry}
\usepackage{lineno}
\usepackage{authblk}

\usepackage{mathpazo}

\usepackage{array}
\usepackage{multirow}

\usepackage{amssymb}
\usepackage{pifont}

\usepackage[colorinlistoftodos]{todonotes}

\usepackage{wrapfig}

\usepackage[numbers,sort&compress]{natbib}
\usepackage{cite}

\hypersetup{
    colorlinks=true, 
    linkcolor=black,  
    citecolor=blue, 
    urlcolor=blue    
}

\usepackage{amsthm}

\usepackage{titlesec}
\usepackage{titletoc}
\titleclass{\subsubsubsection}{straight}[\subsection]
\newcounter{subsubsubsection}[subsubsection]
\renewcommand\thesubsubsubsection{\thesubsubsection.\arabic{subsubsubsection}}
\titleformat{\subsubsubsection}{\normalfont\normalsize\bfseries}{\thesubsubsubsection}{1em}{}
\titlespacing*{\subsubsubsection}{0pt}{3.25ex plus 1ex minus .2ex}{1.5ex plus .2ex}
\titleformat*{\section}{\fontsize{11}{12}\selectfont\bfseries}
\titleformat*{\subsection}{\fontsize{11}{12}\selectfont\bfseries}
\titleformat*{\subsubsection}{\fontsize{11}{12}\selectfont\bfseries}
\titleformat*{\paragraph}{\fontsize{11}{12}\selectfont\bfseries}
\titleformat*{\subparagraph}{\fontsize{11}{12}\selectfont\bfseries}

\titlecontents{subsubsubsection}
[8em] 
{\small}
{\hyperlink{toc.\thecontentslabel}{\thecontentslabel.} }
{}
{\ \titlerule*[.5pc]{.}\contentspage}

\titlespacing*{\subsubsubsection}{0pt}{3.25ex plus 1ex minus .2ex}{1.5ex plus .2ex}

\usepackage{caption}
\usepackage{microtype}

\usepackage{enumitem}
\setlist{itemsep=0em}

\usepackage{fancyhdr} 
\usepackage{lastpage} 

\usepackage{listings}
\usepackage{xcolor}

\definecolor{codegreen}{rgb}{0,0.6,0}
\definecolor{codegray}{rgb}{0.5,0.5,0.5}
\definecolor{codepurple}{rgb}{0.58,0,0.82}
\definecolor{backcolour}{rgb}{0.95,0.95,0.92}

\lstdefinestyle{mystyle}{
    backgroundcolor=\color{backcolour},   
    commentstyle=\color{codegreen},
    keywordstyle=\color{magenta},
    numberstyle=\tiny\color{codegray},
    stringstyle=\color{codepurple},
    basicstyle=\ttfamily\footnotesize,
    breakatwhitespace=false,         
    breaklines=true,                 
    captionpos=b,                    
    keepspaces=true,                 
    numbers=left,                    
    numbersep=5pt,                  
    showspaces=false,                
    showstringspaces=false,
    showtabs=false,                  
    tabsize=2
}

\usepackage{subfig}

\title{Evaluating Pavement Deterioration Rates Due to Flooding Events Using Explainable AI}
\date{}

\author[1]{\normalsize Lidan Peng}
\author[2]{Lu Gao \thanks{lgao5@central.uh.edu}}
\author[3]{Feng Hong}
\author[4]{Jingran Sun}

\affil[1]{Chongqing Jiaotong University}
\affil[2]{Department of Civil and Environmental Engineering, University of Houston}
\affil[3]{Texas State University}
\affil[4]{Center for Transportation Research, The University of Texas at Austin}

\begin{document}

\maketitle

\section*{Abstract}
Flooding can damage pavement infrastructure significantly, causing both immediate and long-term structural and functional issues. This research investigates how flooding events affect pavement deterioration, specifically focusing on measuring pavement roughness by the International Roughness Index (IRI). To quantify these effects, we utilized 20 years of pavement condition data from TxDOT's PMIS database, which is integrated with flood event data, including duration and spatial extent. Statistical analyses were performed to compare IRI values before and after flooding and to calculate the deterioration rates influenced by flood exposure. Moreover, we applied Explainable Artificial Intelligence (XAI) techniques, such as SHapley Additive exPlanations (SHAP) and Local Interpretable Model-Agnostic Explanations (LIME), to assess the impact of flooding on pavement performance. The results demonstrate that flood-affected pavements experience a more rapid increase in roughness compared to non-flooded sections. These findings emphasize the need for proactive flood mitigation strategies, including improved drainage systems, flood-resistant materials, and preventative maintenance, to enhance pavement resilience in vulnerable regions.

\noindent \textbf{Keywords}: Pavement Deterioration, Flooding, Explainable AI, Pavement Performance, Resilience

\section{Introduction}

Pavement deterioration due to flooding leads to several types of damage, including reduced pavement strength, increased roughness, rutting, and cracking, which can lead to a loss of pavement service life and increased rehabilitation costs \citep{abdollahi2024updaps,khan2014developing, dawson2014anticipating, helali2008importance, cechet2005climate, rollings1991pavement, drdacky2010flood, forsyth1987economic}. The economic impact includes direct repair costs, additional user costs related to delays and accidents, and the need for more frequent rehabilitation \citep{hong2024assessment}. Structurally, flooding and moisture infiltration weaken pavement layers, reduce the resilient modulus of unbound materials, and cause differential movement, leading to longitudinal cracks and heaves \citep{asadi2020computational}. The severity of deterioration depends on factors such as pavement structure, drainage, traffic, flood duration, and the condition of the pavement before flooding \citep{valles2023deterioration}. These impacts underscore the importance of integrating flood-related deterioration risks into pavement management systems to support data-driven maintenance planning and efficient resource allocation \citep{gao2012network,ismail2009overview, kulkarni2003pavement, gao2008robust, haas1978pavement}.

To better understand and mitigate these impacts, researchers have employed various methods, including field observations, experimental studies, numerical modeling, and mechanistic-empirical (M-E) simulations. Some studies develop predictive models based on experimental measurements of pavement responses before and after flooding \citep{asadi2020computational}. Numerical tools are also employed to simulate pavement responses under various conditions, with the goal of creating algorithms that can generate reasonable results from a limited number of input parameters for network-level applications. M-E design tools such as UPDAPS and TxME are used to analyze pavement performance under flooding, by modifying layer moduli to reflect moisture-induced weakening of materials \citep{abdollahi2024updaps}. Machine learning techniques are increasingly used to predict pavement performance after flood events, using historical pavement distress and flood data \citep{shariatfar2022effects}. Studies also explored the economic aspects, such as increased maintenance and rehabilitation costs, and proposed frameworks for cost analysis and resource allocation \citep{matini2022development}. Furthermore, research has been conducted to define and evaluate pavement resilience, with the aim of developing methods to minimize loss and improve design and construction practices \citep{nivedya2020framework}.

Despite these advances, assessing flood-induced pavement damage presents several challenges, including the difficulty in obtaining pre- and post-flood data for a comprehensive analysis \citep{yu2021flooded}. Many damage is not visible on the surface immediately after flooding, making it difficult to assess the extent of the problem \citep{mallick2017combined}. Current mechanistic-empirical (ME) based approaches and climate data modeling may not accurately simulate flooding and extreme precipitation events due to limitations in capturing short-term events and moisture damage \citep{abdollahi2024updaps}. In addition, it is challenging to incorporate all flood-related stressors (flood depth, duration, velocity, debris, and contaminants) into a single model \citep{achebe2021incorporating}. Factors such as varying pavement section lengths and heterogeneous flood data sources add complexity to network-level assessments \citep{shariatfar2022effects}. Moreover, accurately predicting long-term deterioration and costs is difficult due to the complex interaction of factors and the variability in agency policies and priorities. Given these limitations, there is a critical need for more in-situ data collection and research to improve the accuracy of pavement deterioration assessments and enhance resilience planning \citep{hong2024assessment}. This, in turn, supports more informed pavement maintenance optimization by identifying priority segments and aligning interventions with available resources \citep{gao2013management}. Motivated by these challenges, this research integrates explainable artificial intelligence (XAI) techniques, specifically SHAP and LIME, with advanced machine learning models. This approach not only enhances prediction accuracy but also provides transparent insights into the contributions of key variables.

The objective of this study is to assess flood-induced pavement degradation by analyzing 20 years of TxDOT PMIS data, and measuring changes in the International Roughness Index (IRI) before and after flood events. Various regression models are developed and evaluated, with explainable AI methods, SHAP and LIM, applied to highlight the contributions of key factors such as initial pavement condition, truck traffic, and flood exposure. The study focuses on isolating the impacts of flooding by comparing flooded and non-flooded pavement sections within the studied region, aiming to provide actionable insights for data-driven decision-making regarding flood mitigation and maintenance planning.

\section{Literature Review}

Roads play a crucial role in facilitating transportation as critical infrastructure components. However, they are vulnerable to damage caused by natural disasters, such as floods. Research consistently demonstrates that rainfall negatively impacts pavement conditions over various time periods \citep{sultana2016modeling}. Therefore, it is important to understand the effects of floods on roads and pavements and implement appropriate systems and designs to prevent or minimize flood-related damage.

\subsection{Impact of Flooding on Pavement Condition and Performance}
Flooding significantly compromises pavement structure and functionality, exhibiting various deterioration patterns \citep{chen2024evaluation,  oliveri2000estimation, parola1998highway, ten2011choice, van2003segmentation, starke2010urban, hollis1975effect, brody2007rising, verstraeten1999nature, perez2008gpr, gaspard2007impact, brody2008identifying}. The intrusion of water into pavement structures elevates the moisture content within unbound materials, consequently diminishing their modulus and load-bearing capabilities \citep{asadi2020computational}. This saturation process undermines the adhesive and cohesive forces that bind aggregates and asphalt, leading to expedited material degradation, such as asphalt stripping \citep{lu2020impact}. Moreover, excess moisture in the asphalt layer can result in weakening of aggregate-bitumen bonding and can lead to a higher rate of damage accumulation in the asphalt layer \citep{abdollahi2024updaps}. Therefore, controlling moisture infiltration and ensuring proper drainage are critical in preventing moisture-induced raveling in asphalt pavements \citep{do2023engineering}. The subsequent reduction in structural support intensifies deflections under load, predisposing the pavement to premature failure mechanisms such as rutting and fatigue cracking. The degree of deterioration is contingent on the attributes of the flood event, the structural composition of the pavement, and prevailing environmental conditions \citep{lu2020impact}. Furthermore, flood-induced debris can obstruct drainage pathways, exacerbating water retention and prolonging the saturation of pavement layers \citep{abdollahi2024updaps}.

The type of damage caused by flooding can be broadly categorized as either visible or hidden \citep{chen2023case}. Visible damage includes readily apparent issues such as cave-ins or washouts of the pavement. Hidden damage, on the contrary, encompasses the weakening of the pavement's internal strength and stiffness. While less immediately evident, hidden damage precipitates rapid deterioration and potential catastrophic failures. The specific stressors associated with flooding events include flood depth, duration, velocity, debris, and contaminants \citep{abdollahi2024updaps}. Flood depth and duration contribute to increased saturation within pavement layers, while flood velocity can inflict immediate structural compromise. Debris accumulation can impede drainage efficiency, further extending water exposure. The impact of flooding events demonstrates distinct patterns, including delayed effects, jump effects (characterized by sudden performance decline), a combination of jump and delayed effects, and direct failure scenarios \citep{lu2020impact}.

The existing state of a pavement significantly influences its response to flooding \citep{valles2023deterioration}. Pavements initially in good or very good condition experience a sudden decline in quality following a flood. Conversely, pavements already in fair or poor condition do not exhibit a substantial immediate decline but undergo accelerated deterioration compared to non-flooded pavements. Stochastic modeling techniques, such as Monte Carlo simulations within discrete-time Markov chains, can effectively represent these deterioration dynamics \citep{gao2007using}. These models facilitate the assessment of pavement condition changes over time under various flooding scenarios, providing valuable insights for infrastructure management.

Effective strategies are available to reduce the adverse effects of flooding on pavement infrastructure. Incorporating efficient drainage systems is critical in mitigating water infiltration and saturation \citep{abdollahi2024updaps}. Employing durable and flood-resistant materials enhances the pavement's capacity to withstand long water exposure. Adopting robust pavement structures reinforces the pavement's capacity to endure flood-induced stresses \citep{chen2024evaluation}. Regular inspection, maintenance, and repair protocols are essential for proactively identifying and addressing vulnerabilities. Implementing comprehensive flood management strategies and actively monitoring weather patterns are also crucial for preparedness and responsiveness during flood events \citep{hong2024assessment}. Innovative solutions, such as integrating geotextile-wrapped sand layers, offer enhanced drainage capabilities and effectively mitigate subgrade moisture fluctuations and swelling pressures \citep{chen2023case}.

\subsection{Deterioration Rates Estimation}
Estimating deterioration rates resulting from flooding requires a comprehensive analysis of the impact of flood events on the road network's condition over time. This can be a complex task as the rate of deterioration depends on multiple factors, including the intensity and duration of the flooding, the road and infrastructure type, and the quality of maintenance and repair work. One effective approach to estimating road deterioration rates due to flooding involves utilizing historical data on flood events and their influence on the road network. Researchers have conducted different studies and proposed methods for predicting the extent of road degradation during floods and also emphasized the significance of consistent road condition monitoring for timely maintenance and rehabilitation efforts. 

For instance, \citet{zhang2008pavement} collected data from pavement sections that were submerged due to flood after Hurricane Katrina and the pavement sections that did not experience flooding. Subsequently, they conducted a systematic analysis and compared the deflection, effective structural number, and resilient modulus of both flooded and non-flooded pavement sections to measure the flood's impact. \citet{chen2014effects} conducted an analysis of pavement performance data before and after Hurricanes Katrina and Rita in Louisiana. The authors calculated and compared the international roughness and rutting depth before and after flooding, as well as the difference in pavement performance between flooded and non-flooded road sections. \citet{shamsabadi2014deterioration} considered the impact of snowstorms and flooding events on pavement performance. They created a regression model incorporating various factors to effectively forecast the increase in IRI. The study highlights the importance of predicting post-natural disaster road conditions to make effective road maintenance planning. \citet{sultana2015study} examined the data from both flooded and non-flooded pavement sections before and after a flooding incident. They observed the flooded section experienced a significant decline in structural strength. This study states the need for long-term monitoring and regular testing of the pavement affected by floods to properly assess the impact of floods like rutting and roughness. \citet{khan2017assessment} assumed that IRI value would jump significantly after flooding event and developed a risk-based framework to evaluate the impact of flooding on road conditions. Their findings indicate that well-constructed road with robust structures and high traffic loads, which is maintained to a high standard demonstrated great resilience to flooding. \citet{sultana2018rutting} addressed the impact of flooding events on pavement deterioration and emphasized the necessity of developing new deterioration models specifically for flood-affected pavements, and presented two mechanistic-empirical models for rutting and roughness in flood-affected pavements. 

Moreover, \citet{valles2023deterioration} studied the 2013 Colorado floods by using statistical analysis and stochastic Markov chains with Monte Carlo simulations to quantify deterioration rates and service life loss based on IRI, and their results highlighted accelerated deterioration in flooded pavements. \citet{shariatfar2022effects} employed a machine learning approach with historical distress data and 2016 Louisiana flood maps to predict post-flooding network-level pavement performance, focusing on key distress types such as roughness and cracking. \citet{hong2023evaluation} utilized GIS analysis and a Markov process to estimate reduction in network-level pavement performance due to a 100-year flood in Texas over a 10-year horizon, linking the findings to infrastructure resilience and maintenance planning. 

\subsection{AI Applications in Pavement Deterioration Modeling}

AI and deep learning models are being increasingly applied to various aspects of pavement deterioration modeling, offering advancements in detection, prediction, classification, and data management \citep{choi2019development,lebaku2024deep,gao2024considering}. These technologies are crucial for optimizing maintenance planning, extending pavement lifespan, and reducing road maintenance costs \citep{lee2019development}.

Recurrent Neural Networks (RNNs), particularly Long Short-Term Memory (LSTM) networks and Gated Recurrent Units (GRUs), are promising in predicting time-series data \citep{hosseini2020use,sun2025simulation}. These models learn from historical pavement condition data, such as crack severity, rutting depth, and the international roughness index (IRI), to forecast future pavement condition indices \citep{yu2023pavement}. Studies have shown that RNN-LSTM models outperforms traditional Deep Neural Networks (DNNs) in pavement damage prediction, achieving lower root-mean-square error (RMSE) values \citep{lee2019development}. DNNs are also employed to predict overall evaluation indices like IRI, crack index (CI), and pavement condition index (PCI) with high accuracy, leveraging data from on-site tests and avoiding deviations from linear regression models \citep{guo2023enhancing}. 

In addition to prediction, Convolutional Neural Networks (CNNs) and hybrid models combining CNNs with RNNs (such as CNN-LSTM) are essential for detection and classification tasks. These deep learning models can automatically extract features from raw data, using images captured by vehicle-mounted cameras, Unmanned Aerial Vehicles (UAVs), or satellite imagery, to detect pavement distresses including cracks, potholes, and other surface deformations \citep{chen2022new}. Deep learning techniques often surpass traditional image processing methods in terms of accuracy and efficiency in identifying these damages, as they do not require manual feature extraction \citep{inacio2023low}. AI models can also detect whether maintenance and rehabilitation treatments have been applied to pavement sections by analyzing condition data over time \citep{gao2023deep}. Moreover, advanced AI techniques, such as Graph Neural Networks (GNNs) and Convolutional Graph Neural Networks (ConvGNNs), are being explored to improve deterioration modeling accuracy by incorporating the spatial structure of road networks, such as the condition of neighboring sections \citep{gao2024considering}. 


\subsection{Research Gap}

Previous studies have relied on mechanistic-empirical models, numerical simulations, and statistical analyses to exam the impact of flooding on pavement deterioration, but these efforts often face significant limitations. First, many are constrained to a single flood event or a short time frame, restricting their ability to capture long-term deterioration trends or the cumulative effects of multiple flooding incidents across a broader network. These studies often depend on small historical datasets tied to a single flood event, limiting the temporal window and geographic scope, thus reducing the generalizability of the findings to larger networks or diverse flood scenarios. Even studies adopting a network-level perspective or over extended periods tend to focus on probabilistic reductions in performance rather than directly estimating deterioration in flooded pavements, often failing to provide specific insights into deterioration rates or mechanisms affecting individual pavement sections.


To overcome these shortcomings, this study analyzes 20 years of pavement condition data from TxDOT’s PMIS database, which covers the entire Texas road network, to provide a comprehensive assessment of flood-induced pavement deterioration across multiple events at a statewide scale. This extensive temporal and geographic scope directly addresses the problem of limited time frames and narrow spatial coverage in prior research. Furthermore, this research integrates explainable AI (XAI) techniques, including SHAP and LIME, into modeling pavement deterioration caused by flooding. While traditional black-box machine learning models offer accurate predictions, they often lack explainability, making it challenging to understand the factors driving pavement degradation. By incorporating XAI, this study enables the identification of key contributors to pavement deterioration at both the aggregate and individual levels.




\section{Methodology}

In recent years, interpretability has gained increasing importance in machine learning and data science. Complex models such as deep neural networks or ensemble methods (e.g., random forests, gradient boosted trees) have achieved state-of-the-art performance across various domains but often lack transparency. Model-agnostic interpretability methods are designed to address this shortcoming without making assumptions specific to a particular type of model. Two popular methods in this category are Local Interpretable Model-Agnostic Explanations (LIME) \citep{ribeiro2016should} and SHapley Additive exPlanations (SHAP) \citep{lundberg2017unified}. LIME aims to explain the model's predictions by approximating it locally around a specific instance. SHAP, on the other hand, leverages game-theoretic concepts, specifically Shapley values, to attribute feature importance in a more globally consistent manner. In this section, we detail the underlying mathematical foundations of LIME and the Shapley value formulation in SHAP. We also discuss how to interpret the outputs of this methodology in a statistically coherent way.

\subsection{LIME}

LIME explains the prediction of a complex model $f$ at a specific instance $\mathbf{x}$ by approximating $f$ locally with a simpler and more interpretable model $g$, often a linear model in the neighborhood of $\mathbf{x}$. The notion of "local" means that LIME focuses on the behavior of $f$ in a small region around the instance $\mathbf{x}$, rather than trying to explain the function $f$ across the entire data distribution \citep{ribeiro2016should}.

Suppose $\mathbf{x}$ is the instance we wish to explain. LIME samples data points in the vicinity of $\mathbf{x}$ (by perturbing features of $\mathbf{x}$) to form a local neighborhood. Denote this neighborhood by $\mathcal{N}(\mathbf{x})$. For each point $\mathbf{z}$ in $\mathcal{N}(\mathbf{x})$, we can compute:
\begin{equation}
\pi_x(\mathbf{z}) = \exp \left( -\frac{D(\mathbf{x}, \mathbf{z})^2}{\sigma^2} \right)    
\end{equation}

\noindent where $D(\mathbf{x}, \mathbf{z})$ is a distance measure (e.g., Euclidean or cosine distance) between the original instance $\mathbf{x}$ and the perturbed instance $\mathbf{z}$. The term $\pi_x(\mathbf{z})$ is often called a weighting kernel, and it assigns higher weights to samples closer to $\mathbf{x}$ \citep{ribeiro2016should}.

Next, LIME fits a simple interpretable model $g$ (e.g., a linear model) in this local neighborhood by solving:
\begin{equation}
\underset{g \in G}{\mathrm{argmin}} \, \sum_{\mathbf{z} \in \mathcal{N}(\mathbf{x})} \pi_x(\mathbf{z}) \bigl( f(\mathbf{z}) - g(\mathbf{z}) \bigr)^2 + \Omega(g).    
\end{equation}

\noindent where, \\
$f(\mathbf{z})$ = the prediction of the original complex model.\\
$g(\mathbf{z})$ = the prediction of the simple interpretable model (often linear in the feature space of $\mathbf{z}$). \\
$G$ = the set of all possible interpretable models under consideration (for instance, linear models). \\
$\Omega(g)$ = a complexity measure for $g$, which ensures that among all possible local approximations, the simpler ones are favored \citep{ribeiro2016should}.

In a linear interpretation, if $g(\mathbf{z}) = w_0 + \sum_j w_j z_j$, the coefficients $w_j$ can be interpreted as the contribution of feature $j$ to the prediction. A higher positive $w_j$ indicates a stronger positive contribution of feature $j$, whereas a negative $w_j$ indicates a negative contribution. The resulting coefficients from the linear model, fitted only to data points near $\mathbf{x}$, provide insight into which features have the greatest impact on the prediction of $f$ in that local region \citep{molnar2020interpretable}. 

\subsection{SHAP}
SHAP is grounded in the concept of Shapley values, which originate from cooperative game theory. In this framework, each feature is regarded as a “player” contributing to the overall prediction of the model. The Shapley value for a given feature quantifies its contribution to the final prediction by considering all possible subsets (coalitions) of other features \citep{lundberg2017unified}. 

Let there be $M$ features in total. For a particular instance $\mathbf{x}$, let the model’s output be denoted by $f(\mathbf{x})$. The idea is to distribute $f(\mathbf{x}) - f(\mathbf{0})$ (the difference between the prediction and some baseline model output) among the $M$ features in a fair manner, where $f(\mathbf{0})$ is often chosen as the output of the model under baseline or reference conditions (e.g., when the features are set to average values) \citep{lundberg2017unified}.

The Shapley value $\phi_j$ for feature $j$ is given by:
\begin{equation}
\phi_j = \sum_{S \subseteq \{1,\ldots, M\} \setminus \{j\}} \frac{|S|! \, (M - |S| - 1)!}{M!} \Bigl[ f \bigl( S \cup \{j\} \bigr) - f(S) \Bigr]    
\end{equation}

\noindent where,\\
$S$ =  any subset of the set of features not including feature $j$.\\
$f(S)$ = the model’s output using the subset $S$ of features. In practice, this can be implemented by marginalizing or in some cases by conditioning on the other features.\\
$\frac{|S|! \, (M - |S| - 1)!}{M!}$  = a weighting factor that ensures each possible ordering of features is counted appropriately \citep{shapley1953value}.

Conceptually, $\phi_j$ is computed by looking at the difference in model output when feature $j$ joins every possible coalition $S$ of other features. This is then averaged over all such subsets of features, giving a comprehensive measure of feature $j$’s importance to the final prediction. The Shapley values $\phi_j$ indicate how much each feature $j$ shifts the model prediction from the baseline $f(\mathbf{0})$ when feature $j$ is introduced into all possible subsets of the other features. A positive Shapley value means that the feature contributes to increasing the prediction relative to the baseline, while a negative value indicates that the feature pulls the prediction below the baseline. Importantly, SHAP values have desirable fairness and consistency properties, making them appealing for interpretability.

\section{Case Study}
\subsection{Data Collection}
In this study, data collection and preprocessing is conducted through a sequence of steps detailed below:

\begin{enumerate}
    \item Information regarding road closures associated with flooding incidents was gathered from TxDOT. More than 100 significant flooding events were identified and evaluated. Example major flooding events include the post-Ike flooding in Southeast Texas in 2008, the flooding event in Montgomery County in 2014, the Lubbock County flooding incident in 2015, and another major flood occurrence in Montgomery County in 2017.
    
    \item The geographic coordinates of both the beginning and ending points of each affected route were obtained via Geographic Information System. These coordinates were subsequently cross-referenced with the Texas Department of Transportation's (TxDOT's) road reference system to identify the respective road reference markers defining the start and end of each route.
    
    \item The relevant pavement sections (each section has a length around 0.5 mile) that fell within the range of the identified beginning and ending points were extracted from TxDOT's Pavement Management Information System (PMIS) \citep{gao2021detection}.
    
    \item Pavement sections that had International Roughness Index (IRI) values available prior to and following the year of the respective flooding event were selectively chosen. In the absence of such information, the sections were considered unsuitable for this analysis and consequently discarded from the dataset.
\end{enumerate}

\begin{figure}[H]
    \centering
    \subfloat[US0180, FM0179, and US0087]{%
        \includegraphics[height=3cm]{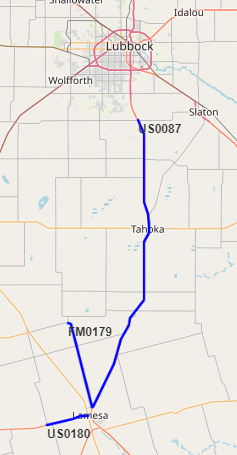}
    }
    \hfill
    \subfloat[FM1908, SH0131, and FM0481]{%
        \includegraphics[height=3cm]{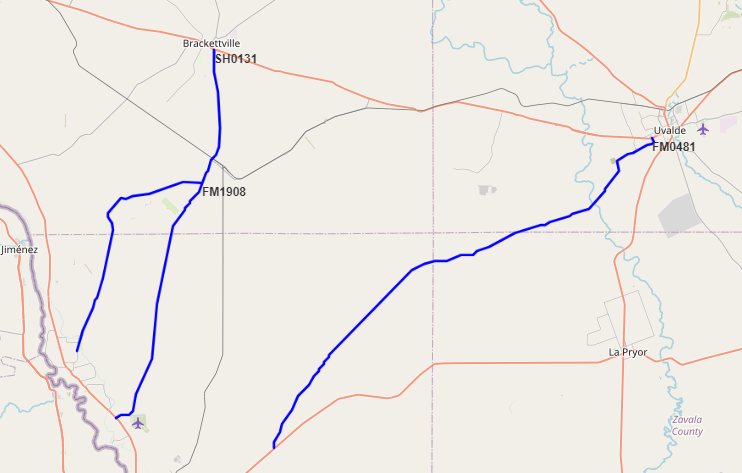}
    }
    \hfill
    \subfloat[FM2854]{%
        \includegraphics[height=3cm]{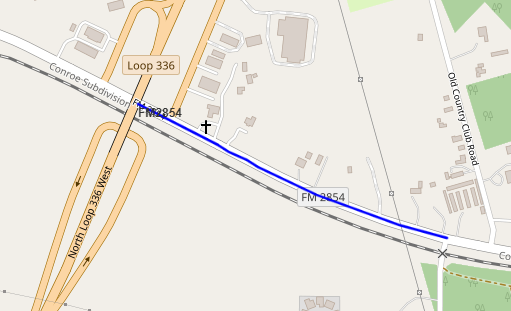}
    }

    \vspace{0.3cm}  

    \subfloat[BU0090Y]{%
        \includegraphics[height=1.8cm]{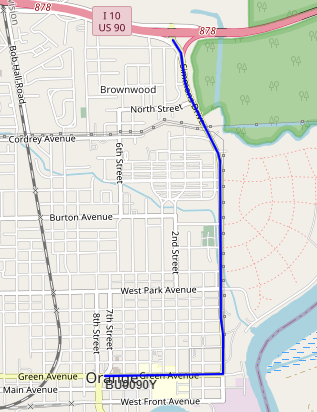}
    }
    \hfill
    \subfloat[FM0366]{%
        \includegraphics[height=1.8cm]{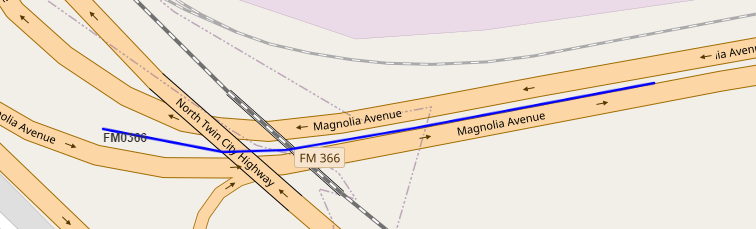}
    }
    \hfill
    \subfloat[FM0597]{%
        \includegraphics[height=1.8cm]{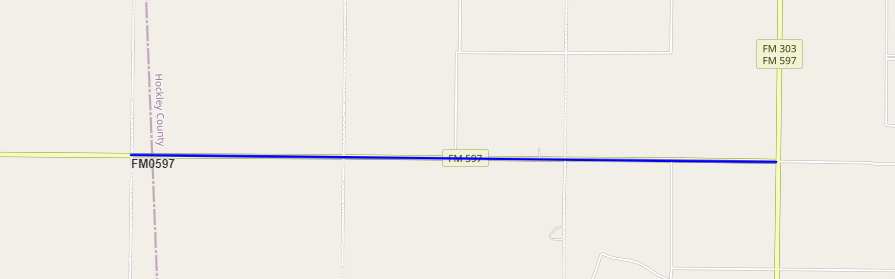}
    }

    \caption{Location of Sample Routes}
    \label{fig:grid}
\end{figure}

Table~\ref{tab:flood_routes} shows sample road closures due to flooding collected in this case study. It includes the route names, flood years, start points, end points, and total lengths (in miles) of the affected road sections. These sample road routes include FM0481 in 2014 (from Eagle Pass to Uvalde, spanning 43.3 miles), FM1908 in 2014 (from Spofford to Quemado, covering 20.6 miles), and SH0131 in 2014 (from Eagle Pass to Brackettville, with a length of 33.1 miles). Figure~\ref{fig:grid} provides a visual representation of these sample locations within the study area.

\begin{table}[H]
    \centering
    \caption{Sample Road Routes Flooded}
    \label{tab:flood_routes}
    \begin{tabular}{lllll l}
        \toprule
        \textbf{Route Name} & \textbf{Flood Year} & \textbf{Start Point} & \textbf{End Point}  \\
        \midrule
        FM0481   & 2014 & Eagle Pass            & Uvalde                  \\
        FM1908   & 2014 & Spofford              & Quemado                \\
        SH0131   & 2014 & Eagle Pass            & Brackettville          \\
        FM2854   & 2017 & Loop 336              & Pinewood Dr.             \\
        US0087   & 2015 & FM 41 south of Lubbock & Lamesa                 \\
        FM0597   & 2015 & Cochran county line   & FM 303                  \\
        FM0179   & 2015 & Lynn county line      & US 87 in Lamesa        \\
        US0180   & 2015 & FM 829                & SH 137                  \\
        FM0366   & 2008 & Intersection with FM 347 & N.A.                  \\
        BU0090Y  & 2008 & FM 1006               & Interstate 10           \\
        \bottomrule
    \end{tabular}
\end{table}





\subsection{Data Preprocessing}

The TxDOT PMIS is a dataset containing comprehensive pavement condition inventory that includes various pavement performance indicators such as potholes, cracking, and rutting. However, not all features are available for every road segment. The IRI is the only condition indicator available for all road segments and is used as the primary indicator in this case study. Moreover, we selected other variables that represent traffic and environmental factors to provide a more comprehensive analysis. The key features included in the dataset are: 

\begin{itemize}
    \item \textbf{TX\_CONDITION\_SCORE}: Pavement condition score (0-100). 100 represents the best score and 0 represents the worst score.
    \item \textbf{TX\_DISTRESS\_SCORE}: Pavement distress score (0-100).
    \item \textbf{TX\_IRI\_AVERAGE\_SCORE}: International Roughness Index (IRI) average score.
    \item \textbf{TX\_TRUCK\_AADT\_PCT}: Percentage of daily traffic that are trucks.
    \item \textbf{TX\_CURRENT\_18KIP\_MEAS}: A measurement related to 18KIP equivalent single-axle loads.
    \item \textbf{TX\_PVMNT\_TYPE\_DTL\_RD\_LIFE\_CODE}: The variable represents the predominant travel lane pavement type during the data collection year for a given section, derived from RLS pavement layer information. It is stored as a string with a maximum length of 100 characters. The possible values include: 01 - Continuously Reinforced Concrete (CRCP), 02 - Jointed Reinforced Concrete (JRCP), 03 - Jointed Plain Concrete (JPCP), 04 - Thick Asphaltic Concrete (greater than 5.5"), 05 - Medium Thickness Asphaltic Concrete (2.5–5.5"), 06 - Thin Asphaltic Concrete (less than 2.5"), 07 - Composite (asphalt-surfaced concrete or ACP on top of a heavily stabilized base), 08 - Widened Composite Pavement, and 09 - Overlaid and Widened Asphaltic Concrete Pavement.
    \item \textbf{CLIMATE\_ZONES}: Climate zones were defined based on temperature and precipitation data obtained from the National Oceanic and Atmospheric Administration (NOAA) database. A 30-year annual average of temperature and precipitation was used to represent the climate. For counties without a weather station, data from adjacent counties were averaged, while for counties with multiple weather stations, the average of those stations was used. Counties were grouped into climate zones corresponding to the west, east, north, south, and central regions.
    \item \textbf{TX\_RURAL\_URBAN\_CODE}: Categorical code indicating if a road section is rural or urban.
    \item \textbf{Flood}: Binary indicator (1 if the road segment recorded as flooded that year, 0 otherwise).
\end{itemize}

The target variable is next year's TX\_IRI\_AVERAGE\_SCORE. 10,022 data records were collected for this case study. Each data point represents a pavement section of length around 0.5 mile. Table \ref{tab:descriptive_statistics} summarizes the descriptive statistics of the numerical variables. This includes the mean, standard deviation, minimum, maximum, and quartiles. Among the 10,022 pavement sections, 501 (5\%) were classified as flooded, while the remaining 9,521 (95\%) were identified as non-flooded. The data points were selected based on the availability of key features. Pavement sections with missing values for the features listed in Table \ref{tab:descriptive_statistics} were excluded from the statistical analysis.


\begin{table}[H]
\centering
\caption{Descriptive Statistics of the Numerical Features}
\label{tab:descriptive_statistics}
\begin{tabular}{lrrrrr}
\toprule
\textbf{Feature} & \textbf{Mean} & \textbf{Std. Dev.} & \textbf{Min} & \textbf{25\%} & \textbf{Max} \\
\midrule
TX\_CONDITION\_SCORE        & 93.91  & 13.87  &  0.00  & 97.00  & 100.00 \\
TX\_DISTRESS\_SCORE         & 95.70  & 11.35  &  0.00  & 99.00  & 100.00 \\
TX\_IRI\_AVERAGE\_SCORE     & 100.61 & 54.17  & 26.00  & 57.00  & 314.00 \\
TX\_TRUCK\_AADT\_PCT        & 17.60  &  8.52  &  0.00  & 13.10  &  56.90 \\
TX\_CURRENT\_18KIP\_MEAS    & 1096.57 & 978.45  &  0.00  & 200.00  & 8123.00 \\
TX\_PVMNT\_TYPE\_DTL\_RD\_LIFE\_CODE & 8.74  & 1.98  & 1.00  & 6.00  & 10.00 \\
CLIMATE\_ZONES\_encoded    & 1.83  & 0.98  & 0.00  & 2.00  & 3.00 \\
TX\_RURAL\_URBAN\_CODE     & 1.03  & 0.21  & 1.00  & 1.00  & 4.00 \\
Flood                      & 0.05  & 0.21  & 0.00  & 0.00  & 1.00 \\
\bottomrule
\end{tabular}
\end{table}



\begin{figure}[H]
\centering
\includegraphics[width=0.85\textwidth]{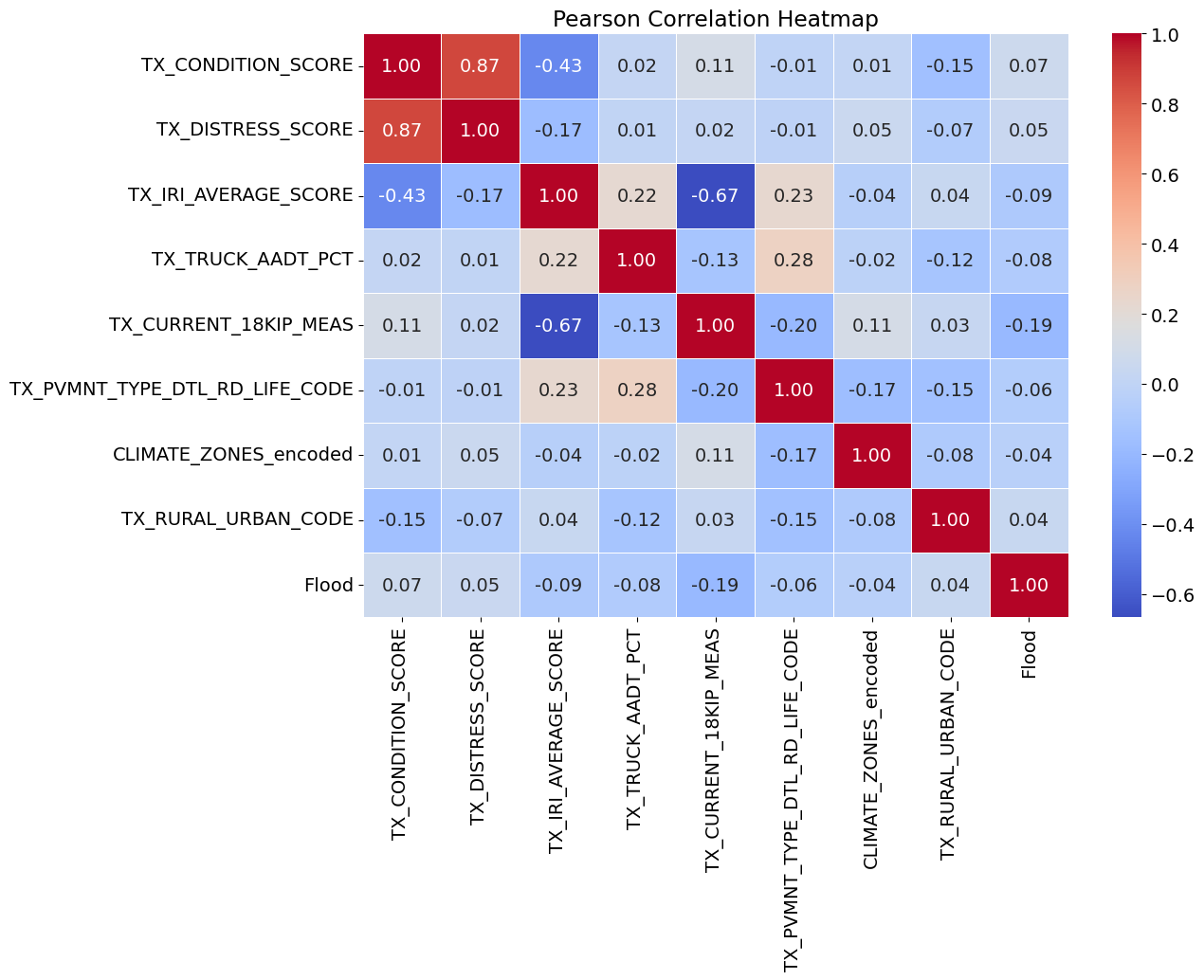} 
\caption{Correlation matrix of selected numerical features.}
\label{fig:correlation_matrix}
\end{figure}

Figure \ref{fig:correlation_matrix} presents the Pearson correlation heatmap for features used in this case study. A high positive correlation (0.87) is observed between TX\_CONDITION\_SCORE and TX\_DISTRESS\_SCORE, which indicates that the overall pavement condition rating is strongly influenced by the distress score. TX\_CONDITION\_SCORE shows a moderate negative correlation (-0.43) with TX\_IRI\_AVERAGE\_SCORE. The negative relationship arises because the IRI value indicates worse pavement roughness as it increases, whereas a higher pavement condition score indicates better pavement condition. Traffic-related variables, such as TX\_TRUCK\_AADT\_PCT, show weak correlations with most pavement condition indicators, with the highest correlation (0.22) observed with TX\_IRI\_AVERAGE\_SCORE. The correlation between Flood and current pavement condition variables is relatively weak. 

\subsection{Statistical Analysis}
\subsubsection{Before and After Flooding IRI Comparison}
\begin{figure}[H]
    \centering
    \includegraphics[width=0.75\linewidth]{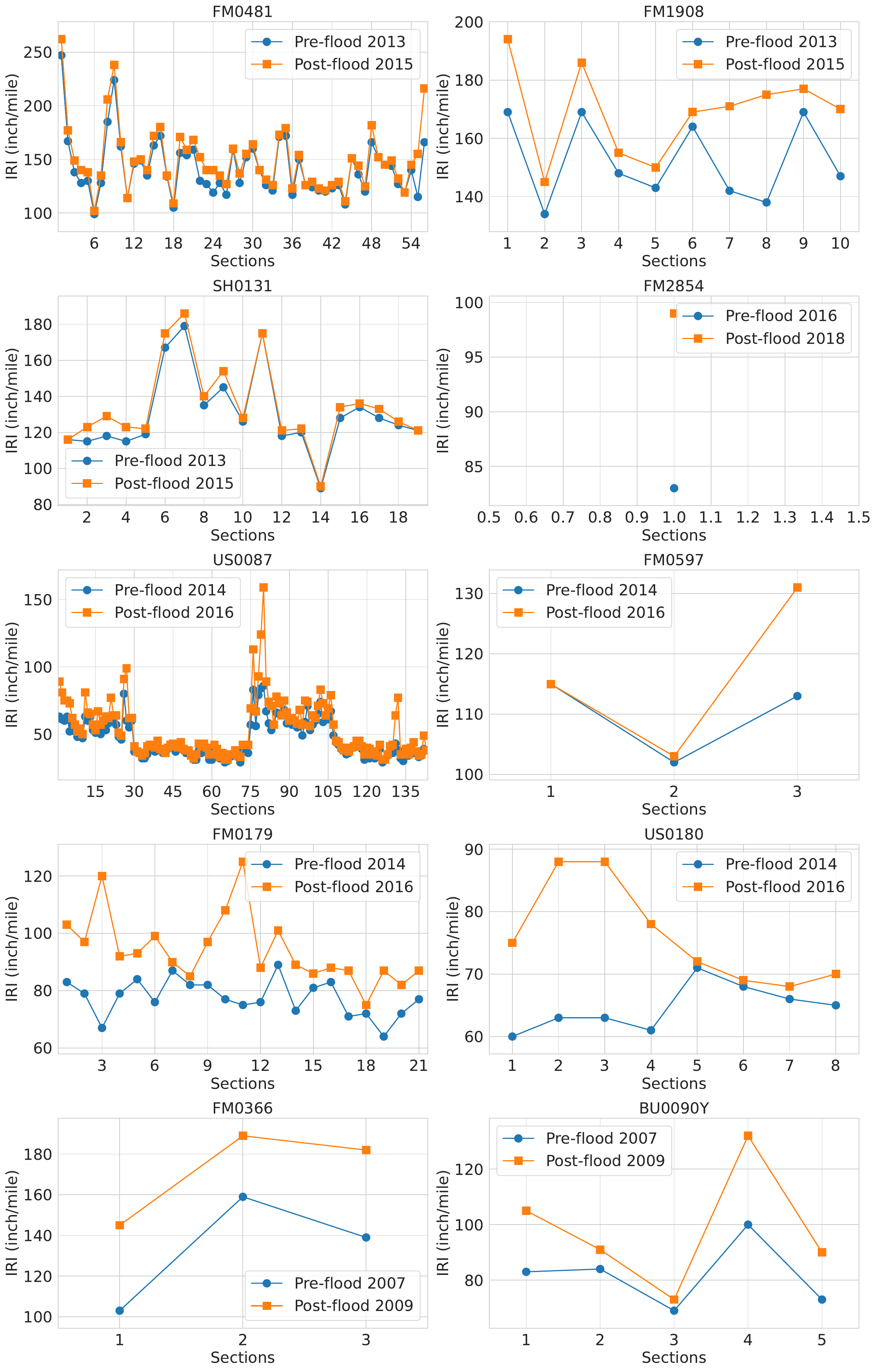}
    \caption{IRI before and after the flooding events}
    \label{fig:pre_post}
\end{figure}

Figure \ref{fig:pre_post} illustrates the changes in IRI values, measured in inches per mile (in/mi), across multiple roadway routes before and after flooding events. Each subplot corresponds to a specific route (e.g. FMD481, SH0131, US0087) and displays pre-flood (one year prior to flooding) and post-flood (one year after flooding) IRI measurements for individual sections within the route. For instance, route FMD481 (pre-flood 2013, post-flood 2015) shows the post-flood IRI value of one section reaching up to 200 in/mi, significantly higher than the pre-flood level. Similar trends are evident in other routes, such as US0087 and FM0179, where post-flood IRI values rise sharply.

While the overall trend confirms higher (worse) post-flood IRI values, variability exists among sections. For example, SH0131 (pre-flood 2013, post-flood 2015) exhibits minimal deterioration in some sections. This inconsistency may result from localized flood impacts, temporary route closures by TxDOT during floods, or post-flood maintenance interventions that restored certain segments.

To ensure data reliability, sections lacking complete IRI records during the two-year analysis window (one year before and after flood) were excluded. Also, sections with improved IRI values (potentially due to maintenance) were omitted to isolate flood-induced deterioration. In pavement performance modeling, when maintenance records are unavailable, it is common to exclude pavement condition data that show improvement over consecutive years \citep{mishalani2002computation}. This is because pavement conditions are not expected to improve without maintenance intervention. In this case study, maintenance records for the analyzed road segments are unavailable. Therefore, if IRI values of a road segment decrease after flooding, it is assumed that maintenance was performed, and these data points are excluded from the statistical analysis.

The analysis reveals an average IRI increase of 8.46 in/mi over two years, with a high standard deviation (10.53 in/mi), indicating significant variability in deterioration severity across sections. This underscores that while flooding generally exacerbates pavement roughness, the extent of damage is highly dependent on localized factors such as flood intensity, road segment vulnerability, and maintenance responsiveness.

\subsubsection{Before and After Flooding IRI Deterioration Rate Comparison}
\begin{figure}[H]
    \centering
    \includegraphics[width=1\linewidth]{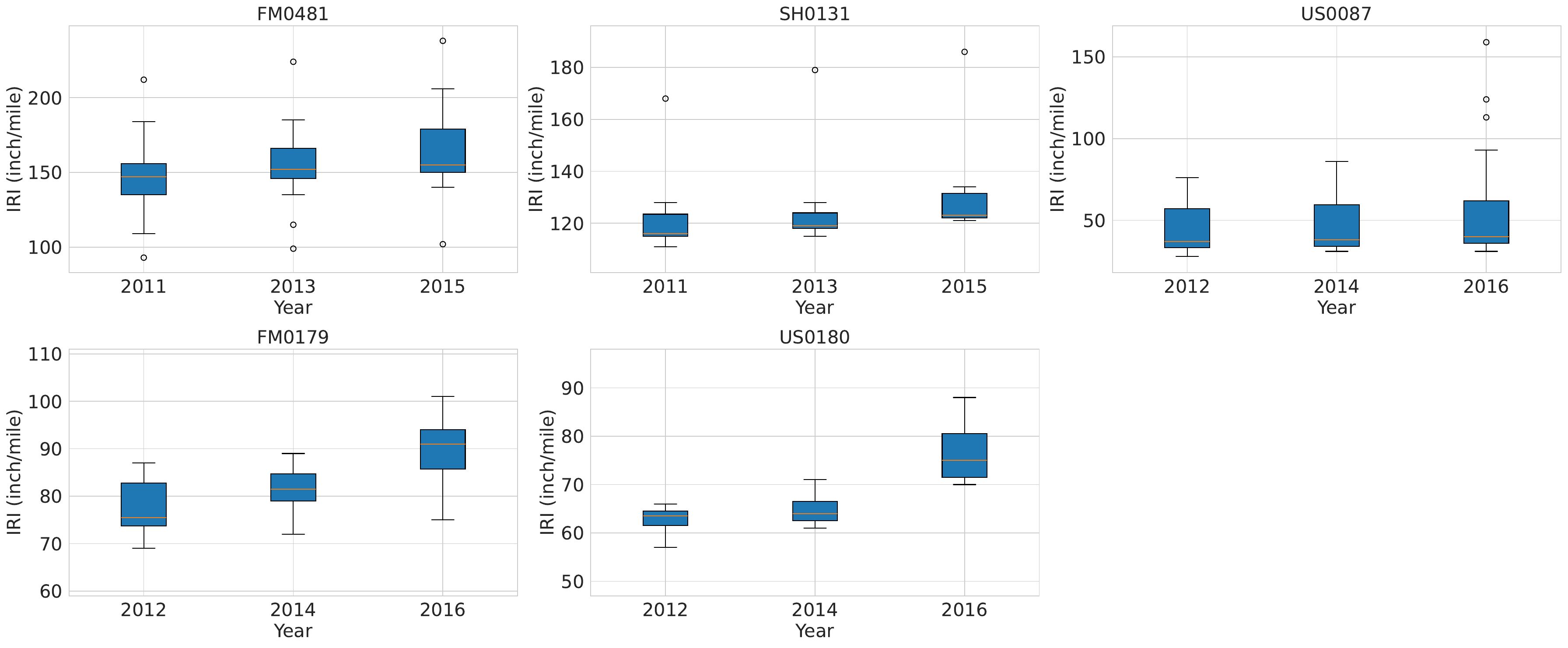}
    \caption{IRI Before and After Flooding Events}
    \label{fig:trend}
\end{figure}

Figure \ref{fig:trend} illustrates the IRI values during three different years: three years prior to flooding, one year before flooding, and one year after flooding. The chosen interval of two years allows for a comparison of the deterioration rate before and after the flooding event. One-year intervals were not used due to the lack of precise data collection dates for the IRI values. Consequently, it is difficult to determine whether the IRI at the time of flooding was measured before or after the event occurred. The IRI values presented in Figure \ref{fig:trend} represent the average values across sections in the same route. When calculating the average, sections lacking IRI data in these three years were excluded. Furthermore, sections where the IRI values improved, indicating potential maintenance effects, were also excluded. For this reason, not all routes were plotted in the figure.  The figure shows that, in general, IRI deterioration rate increases after flooding. However, the rate of deterioration varies across different segments. Some sections experience a much sharper decline in condition, as seen in certain data points for US0087, where roughness increases significantly post-flooding. In contrast, there are cases like SH0131, where the impact of flooding appears minimal, with IRI values remaining relatively stable.


\subsubsection{Flooded and Non-Flooded Sections Comparison}
\begin{figure}[H]
    \centering
    \includegraphics[width=0.75\linewidth]{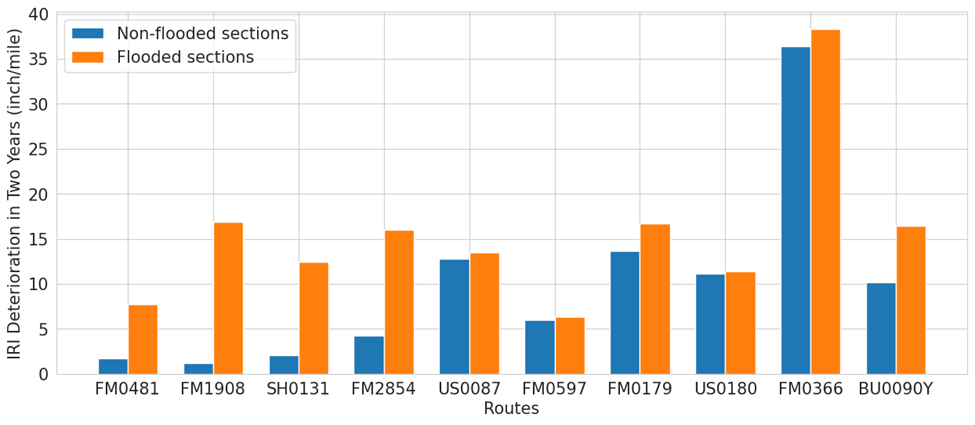}
    \caption{IRI Deterioration Rates: Flooded vs. Non-Flooded Sections}
    \label{fig:non_flood}
\end{figure}

Figure \ref{fig:non_flood} shows the comparison of the IRI deterioration rate between sections affected by flooding and sections outside the flooding region nearby. In order to compare them under the same condition, both flooded and non-flooded sections are collected from the same route. The x-axis represents different sections and the y-axis represents the IRI difference between the year after flooding and the year before flooding, representing the IRI deterioration in two years. As can be seen from Figure \ref{fig:non_flood}, the differences between the flooded and non-flooded sections are range from 0.3 to 15 inch/mile. The average IRI change difference between flooded and non-flooded sections is 5.6 inch/mile. 

\subsection{XAI Modeling}


\subsubsection{Machine Learning Models}

A total of six regression models were trained and evaluated to predict next year's IRI: Linear Regression (LR), Ridge Regression (Ridge), Lasso Regression (Lasso), Decision Tree (DT), Random Forest (RF), and Gradient Boosting Regression (GBR). The dataset was divided into training (80\%) and test (20\%) subsets through random sampling to ensure a balanced representation of pavement conditions and external factors. 

We implemented all models using the scikit-learn package and initially used default settings. Hyperparameter tuning was performed to optimize performance for certain models. For Ridge and Lasso models, we adjusted the regularization strength ($\alpha$) using a grid search over values from $10^{-3}$ to $10^2$. The DT model was optimized by tuning the maximum tree depth ($2-20$), the minimum samples required to split a node (2–10), and the minimum samples per leaf (1–5). RF tuning included adjusting the number of trees (50, 100, 200), maximum depth (5–20), and minimum samples per split (2, 5, 10), along with feature subsampling. GBR was optimized by adjusting the learning rate (0.001, 0.01, 0.1), number of estimators (100, 300, 500), maximum tree depth (3, 5, 10), and subsampling ratios (0.5, 0.75, 1.0). The best hyperparameters were determined using 5-fold cross-validation on the training set, selecting the combination that minimized mean squared error (MSE). 


Each of the six regression models was evaluated using mean squared error (MSE), mean absolute error (MAE) and $R^2$ score on the test set. Table~\ref{tab:model_performance} summarizes the performance. Among the models evaluated, GBR demonstrated the best performance, achieving the lowest mean squared error (MSE) of 247.0450, the lowest mean absolute error (MAE) of 8.6095, and the highest $R^2$ score of 0.9310. This result suggests that GBR effectively captures complex relationships within the dataset. The LR, Ridge, and Lasso exhibited nearly identical performance, with MSE values around 263 and $R^2$ scores of approximately 0.9263–0.9266. Lasso Regression achieved a slightly better MSE (262.8292) and $R^2$ score (0.9266) compared to the standard Linear Regression model, though the improvement was marginal.  The DT model performed significantly worse than the other models, with the highest MSE (360.7363) and MAE (10.6944), resulting in a notably lower $R^2$ score of 0.8993. The RF model achieved slightly better performance than the DT, with an MSE of 264.0790 and an $R^2$ score of 0.9262. Since GBR outperformed all other models in terms of predictive accuracy, we use it for further XAI analysis to better understand the contributions of different features to pavement deterioration.

\begin{table}[H]
\centering
\caption{Comparison of Regression Models for Predicting Next Year's IRI}
\label{tab:model_performance}
\begin{tabular}{lccc}
\toprule
\textbf{Model} & \textbf{MSE} & \textbf{MAE} & \textbf{R$^2$ Score} \\
\midrule
Linear Regression (LR)  & 263.8513 & 8.8476 & 0.9263 \\
Ridge Regression (Ridge)  & 263.7546 & 8.8463 & 0.9263 \\
Lasso Regression (Lasso)  & 262.8292 & 8.8517 & 0.9266 \\
Decision Tree (DT)       & 360.7363 & 10.6944 & 0.8993 \\
Random Forest (RF)       & 264.0790 & 8.5476 & 0.9262 \\
Gradient Boosting (GBR)  & \textbf{247.0450} & \textbf{8.6095} & \textbf{0.9310} \\
\bottomrule
\end{tabular}
\end{table}

\subsubsection{Feature Importance via SHAP}
\begin{figure}[H]
\centering
\includegraphics[width=0.9\textwidth]{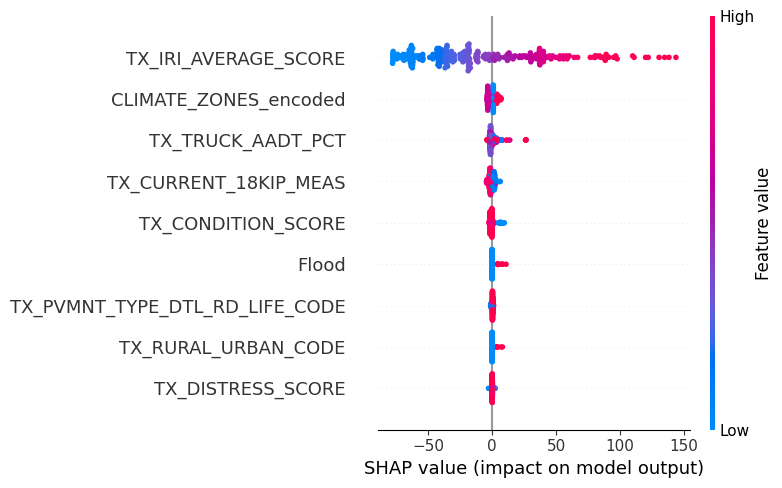} 
\caption{SHAP summary plot for the gradient boosting model.}
\label{fig:shap_summary}
\end{figure}

SHAP values were computed to assess the global contribution of each feature to the model's predictions. Figure \ref{fig:shap_summary} shows a summary plot of the SHAP values for all features. The SHAP summary plot provides insights into how key variables influence the predictions of a GBR model, which is trained to predict next year’s IRI. The features are ranked by their mean absolute SHAP values, with the most influential variables appearing at the top. The results indicate that TX\_IRI\_AVERAGE\_SCORE is the most critical predictor, followed by CLIMATE\_ZONES, TX\_TRUCK\_AADT\_PCT, and TX\_CURRENT\_18KIP\_MEAS. 

The SHAP values along the x-axis indicate how each feature affects the model's predictions. Positive SHAP values increase the predicted IRI, meaning worse pavement conditions, while negative SHAP values decrease the predicted IRI, indicating better pavement performance.  The color coding in the SHAP plot represents the feature values, where red denotes high values and blue denotes low values. 

TX\_IRI\_AVERAGE\_SCORE has the greatest impact on next year's IRI value. Higher current IRI scores (red) significantly increase the predicted IRI, suggesting that roads already rough currently are likely to become even rougher next year. Lower current IRI scores (blue) tend to have a neutral or slightly negative impact, indicating that smoother roads are likely to remain relatively smooth or improve slightly. TX\_TRUCK\_AADT\_PCT has a moderate impact. Higher percentages of truck traffic (red) increase predicted IRI, indicating that roads with more truck traffic are likely to become rougher due to the heavy loads accelerating pavement deterioration. Lower percentages (blue) have a neutral or slightly negative impact, suggesting that roads with less truck traffic are likely to experience a less increase in roughness or maintain smoother surfaces. For TX\_CURRENT\_18KIP\_MEAS, the balanced distribution and moderate spread of SHAP values indicate that its influence varies depending on the specific load magnitude across data instances. Low values of TX\_CURRENT\_18KIP\_MEAS (blue) are associated with higher SHAP values, indicating that weaker pavements deteriorate faster. 
    
The Flood feature also shows a tendency toward positive SHAP values, suggesting that flood-prone areas experience higher predicted IRI values, indicating accelerated pavement deterioration. This outcome aligns with the physical effects of flooding on roadways. The SHAP distribution of the Flood variable indicates that flood-exposed roadways are more likely to suffer from worsening pavement conditions in the following year. The effect of flooding is evident, as instances with high flood exposure show increased SHAP values, indicating that the role of extreme weather events in pavement degradation.

\subsubsection{Local Interpretations via LIME}

\begin{figure}[H]
\centering
\includegraphics[width=0.95\textwidth]{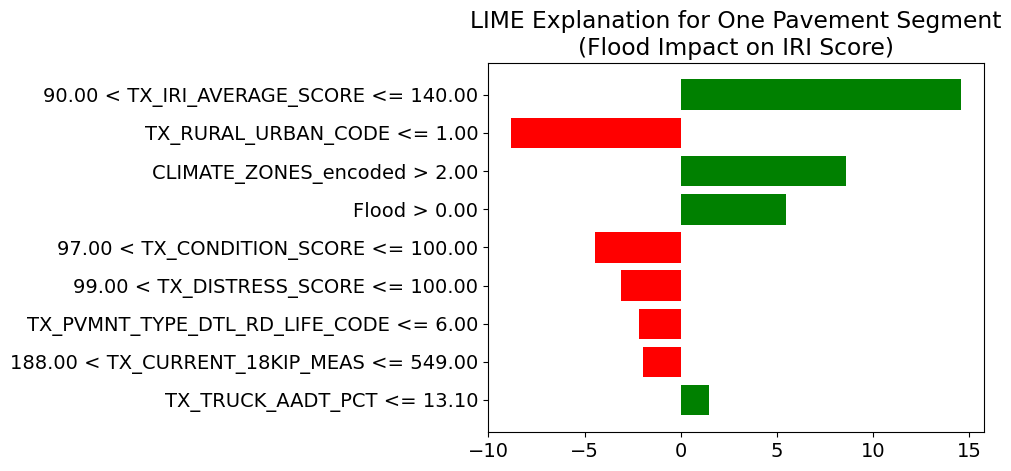} 
\caption{Example LIME explanation for one pavement segment, indicating that flood occurrence contributes to higher IRI (indicating worse condition).}
\label{fig:lime_explanation}
\end{figure}

Figure~\ref{fig:lime_explanation} presents a LIME analysis for one pavement segment and the impact of individual features on the predicted IRI. The bars represent the direction and magnitude of each feature's contribution to the model’s prediction, where green bars indicate a positive impact (leading to a higher predicted IRI, meaning worse pavement condition) and red bars indicate a negative impact (leading to a lower predicted IRI, meaning better pavement condition). The most influential feature in this instance is TX\_IRI\_AVERAGE\_SCORE, which falls within the range of 90.00 to 100.00. Its green bar extending farthest to the right indicates that a high initial roughness significantly contributes to a higher predicted IRI value. The Flood feature, with a condition of \textit{Flood > 0.0}, appears with a green bar positioned around +5 on the x-axis, indicating that flood occurrence contributes to an increase in IRI by approximately 5 units. This aligns with the expectation that flooding accelerates pavement deterioration. The moderate impact suggests that while flooding plays a role in roughness progression, other factors such as initial pavement condition may have a greater influence in this particular instance. TX\_CONDITION\_SCORE has a negative impact on IRI, which is expected since higher condition scores indicate better pavement quality and are associated with lower roughness. Similarly, TX\_DISTRESS\_SCORE follows the same trend: higher distress scores in this range correlate with lower IRI values. Moreover, TX\_CURRENT\_18KIP\_MEAS has a minor negative impact, indicating that higher axle loads are linked to reduced roughness in this particular case. This could be due to the presence of more robust pavement structures on roads designed to accommodate heavy traffic, leading to better surface conditions. TX\_TRUCK\_AADT\_PCT, representing truck traffic percentage, has a red bar, implying that in this specific case, the given level of truck traffic is associated with a lower predicted IRI. This may be due to well-maintained roads in high-traffic areas or the presence of more durable pavement designs. Overall, Figure~\ref{fig:lime_explanation} provides a localized interpretation of how various factors influence the IRI prediction for one specific pavement segment. It confirms that flood exposure significantly contribute to pavement deterioration, while some features, such as  truck traffic percentage, may correlate with reduced roughness in this specific case. 

\subsection{Discussion}






This research provides important insights into the impact of flooding on pavement deterioration. The analysis reveals an average IRI increase of 8.46 in/mi over two years, with a high standard deviation of 10.53 in/mi, indicating significant variability in deterioration severity across sections. While flooding generally exacerbates pavement roughness, the extent of damage largely depends on localized factors such as flood intensity, road segment vulnerability, and maintenance responsiveness. On average, the deterioration rate of IRI increased compared to the same two-year period before the flooding, which highlights the substantial impact of the flood event. Moreover, the difference in IRI changes between flooded and non-flooded sections ranges from 0.3 to 15 in/mi, with an average difference of 5.6 in/mi. These findings align with previous studies, such as \citet{shariatfar2022effects}, who observed a slightly higher rate of IRI increase in flooded areas after flood. Similarly, \citet{valles2023deterioration} reported a sudden drop in the Riding Index (RIDX) of 27 and 11 points for pavements in very good and good condition, respectively, corresponding to a rise in IRI and a quantified service life reduction of 14.3\% for pavements in good condition, suggesting an accelerated deterioration rate due to flooding.

The application of SHAP analysis highlights that flooding, along with truck traffic percentage and initial pavement condition, are among the most influential factors contributing to pavement degradation. Moreover, LIME explanations confirm that flood exposure significantly contributes to increased roughness. The integration of XAI techniques into pavement deterioration analysis not only enhances model transparency but also facilitates data-driven decision-making for transportation agencies. SHAP analysis allows for a global understanding of feature importance. On the other hand, LIME provides local interpretability, helping identify specific instances where unexpected deterioration trends occur. The combination of these techniques can bridge the gap between traditional black-box machine learning models and actionable insights. 

The findings from this study can guide transportation agencies in prioritizing maintenance for flood-prone roads by focusing on sections with higher IRI increases and variability. Proactive scheduling of preventive maintenance, such as resurfacing and drainage improvements, can help extend the service life of pavement and prevent severe deterioration. Moreover, applying flood-resistant materials and improved drainage systems in high-risk areas can enhance infrastructure resilience.

\section{Conclusions}

This study evaluates the impact of flooding on pavement deterioration using 20 years of pavement condition data from TxDOT’s PMIS database, focusing on roughness changes measured by IRI. The findings indicate that flood-affected pavement sections experience  significantly faster deterioration than non-flooded roads. On average, the IRI increased by 8.46 inches/mile over two years for flooded sections. Moreover, the deterioration rate after flooding was significantly higher than the rate observed in the two years before flooding. When comparing flooded and non-flooded pavement sections, the study found that flooded roads deteriorated by an additional 5.6 inches/mile on average, with some sections even experiencing an extreme deterioration difference of up to 15 inches/mile. To understand the underlying causes of pavement degradation, this research applied Explainable AI (XAI) techniques, including SHAP and LIME. The results identified initial pavement roughness (IRI values), truck traffic percentage, flood exposure, and climate zone conditions as the most influential factors affecting pavement deterioration. These findings help transportation agencies focus maintenance efforts on flood-prone roads with high IRI increases and deterioration variability.

This research has some limitations that should be acknowledged. First, the study is based on a limited number of flooding events in the case study, which may restrict the generalization of our findings. To enhance the robustness of our conclusions, it would be beneficial to include a more extensive range of flooding events with more diverse geographic locations in future studies. Second, the availability of data points posed a constraint on our research. As we only had access to PMIS data up to 2018, some recent flooding events had to be excluded from this study. Moving forward, efforts will be made to address these two limitations and expand the scope of the research to encompass a broader set of events and more up-to-date data. Third, maintenance activities that might have influenced post-flood IRI values could not be entirely accounted for, potentially affecting the precision of deterioration estimates. Future research should aim to incorporate past maintenance work records. Fourth, while various pavement performance indicators could provide additional insights into the impact of flooding, these parameters (e.g., cracking, rutting, and raveling) were either not available in the dataset or were too incomplete to allow for statistical analysis in this study. Although data on potholes, cracking, and rutting were present, they covered only a small fraction of the analyzed sections. In contrast, IRI values were available for all sections, which is the reason why IRI was selected as the primary indicator. Future studies should explore datasets with more comprehensive pavement condition metrics to enhance the analysis.

\bibliographystyle{unsrtnat}
\bibliography{references}

\end{document}